\PassOptionsToPackage{dvipsnames}{xcolor}
\documentclass[10pt, a4paper]{article}

\usepackage{lrec-coling2024}\usepackage{latexsym}
\usepackage{amsthm}
\usepackage{float}
\usepackage{subfig}
\usepackage{bm}
\usepackage{amsmath}
\usepackage{amsfonts}
\usepackage{booktabs}
\usepackage{multirow}

\usepackage[dvipsnames]{xcolor}\definecolor{mygreen}{HTML}{B9E0A5}
\definecolor{myred}{HTML}{F19C99}

\title{
A Cause-Effect Look at Alleviating Hallucination of \\ Knowledge-grounded Dialogue Generation
}

\name{
Jifan Yu$^\natural$ \ Xiaohan Zhang$^\clubsuit$ \ Yifan Xu$^\flat$$^\sharp$ \ Xuanyu Lei$^\flat$$^\sharp$ \\
\large \bf Zijun Yao$^\flat$$^\sharp$ \quad Jing Zhang$^\spadesuit$ \quad Lei Hou$^\flat$$^\sharp$ \quad Juanzi Li$^\flat$$^\sharp$$^\dagger$
\thanks{
        $^\dagger$ Corresponding author.
    }
} 

\address{
$^\natural$Institute of Education; $^\flat$BNRist; $^\sharp$KIRC, Institute for Artificial Intelligence \\
Tsinghua University, Beijing 100084, China \\
$^\clubsuit$Zhipu.AI; $^\spadesuit$Renmin University \\
\{yujifan,houlei,lijuanzi\}@tsinghua.edu.cn \\
}

\abstract{
Empowered by the large-scale pretrained language models, existing dialogue systems have demonstrated impressive performance conducting fluent and natural-sounding conversations. However, they are still plagued by the \emph{hallucination} problem, causing unpredictable factual errors in the generated responses. Recently, knowledge-grounded dialogue generation models, that intentionally invoke external knowledge resources to more informative responses, are also proven to be effective in reducing hallucination. Following the idea of getting high-quality knowledge, a few efforts have achieved pretty good performance on this issue. As some inevitable knowledge noises may also lead to hallucinations, it is emergent to investigate the reason and future directions for building noise-tolerant methods in KGD tasks. In this paper, we analyze the causal story behind this problem with counterfactual reasoning methods. Based on the causal effect analysis, we propose a possible solution for alleviating the hallucination in KGD by exploiting the dialogue-knowledge interaction. Experimental results of our example implementation show that this method can reduce hallucination without disrupting other dialogue performance, while keeping adaptive to different generation models. We hope our efforts can support and call for more attention to developing lightweight techniques towards robust and trusty dialogue systems.
\\ 
\newline 
}

\begin{document}

\maketitleabstract

\section{Introduction}
\label{sec:intro}

\begin{figure*}[t]
    \centering
    \includegraphics[width=1.0\linewidth]{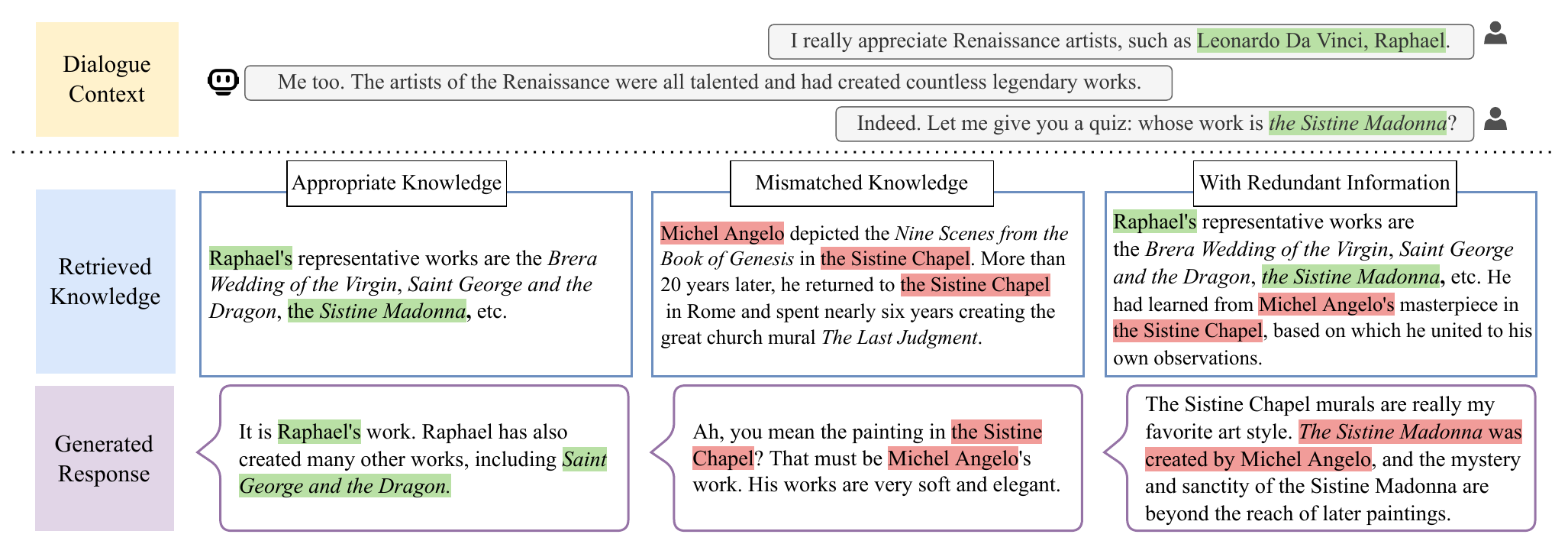}
    \caption{An example of the hallucination of knowledge-grounded dialogue. The figure presents three scenarios where the retrieved knowledge is perfectly matched, mismatched and with some redundant information. The texts with shading are dialogue-relevant knowledge. \sethlcolor{mygreen} \hl{Green} ones means the accurate contents. \sethlcolor{myred} \hl{Red} ones are the possible causes and effects of hallucination.}
    \label{fig:cases}
\end{figure*}

The conversation is an exchange of sentiments and ideas between two or more speakers, from which they grow common ground and subsequently sustain long-term relationships~\cite{clark2019makes}. If a response tends to be unfaithful, \textit{i.e.}, by producing content that conflicts with facts or previous dialogue, it may run the risk of jeopardizing the entire
experience of conversation~\cite{dziri2021neural}. Unfortunately, even the modern dialogue systems, which are powered by pretrained language models~\cite{brown2020language,du-etal-2022-glm}, still can not fully avoid generating responses with self-contradiction or even factual errors. This phenomenon, as the example shown in Figure \ref{fig:cases}, has been generally summarized as the issue of \textit{Knowledge Hallucination}~\cite{dziri2021evaluating,ishii2022survey}, which gradually becomes an emergent challenge toward a reliable, robust AI dialogue system~\cite{marcus2020next}.

In recent years, a series of researches have explored that leveraging external knowledge can help lift the informativeness as well as the factual accuracy~\cite{dinan2018wizard,huang2021plato} of the generated responses. This line of techniques, known as knowledge-grounded dialogue (KGD) generation~\cite{zhou2020kdconv,zheng2021exploring}, generally retrieve the most relevant knowledge with current dialogue and generate an appropriate response based on it. To further alleviate the hallucination problem, one of the straightforward ideas is to guarantee the quality and matching accuracy of the selected external knowledge. Benefited from this thinking, methods such as augmenting knowledge with web retrievers~\cite{shuster2021retrieval} and conducting refinement by re-querying knowledge bases~\cite{dziri2021neural} have achieved pretty good performance in this task.

Despite these advanced methods that attempt to guarantee the quality of retrieved knowledge, there remain some inevitable knowledge noises in many scenarios. First, real-world knowledge sources still face the challenge of incompleteness and incorrectness~\cite{wang2021survey}, which occasionally leads to mismatching issues for KGD models. Second, a correct knowledge piece may preserve redundant contents that can possibly cause subsequent hallucinations. As the example illustrated in Figure \ref{fig:cases}, once the retrieved knowledge is not so relevant to dialogue query, or even contains some extra information, there is a potential risk that the \emph{parametric knowledge bias}~\cite{ishii2022survey} of generation model is activated, which eventually leads to hallucinations in generated responses.

\begin{figure}[t]
	\centering
	\subfloat[]{\label{fig:causal_a}\includegraphics[width=0.38\linewidth]{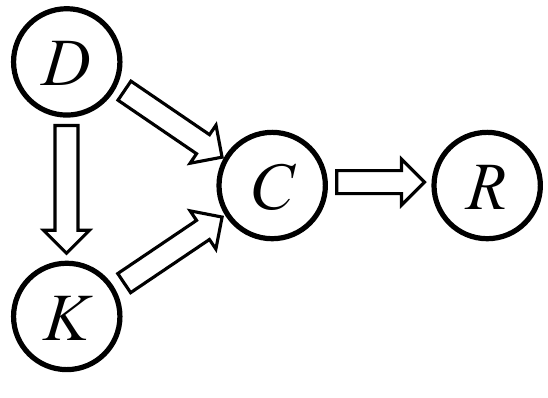}}\\
	\subfloat[]{\label{fig:causal_b}\includegraphics[width=0.8\linewidth]{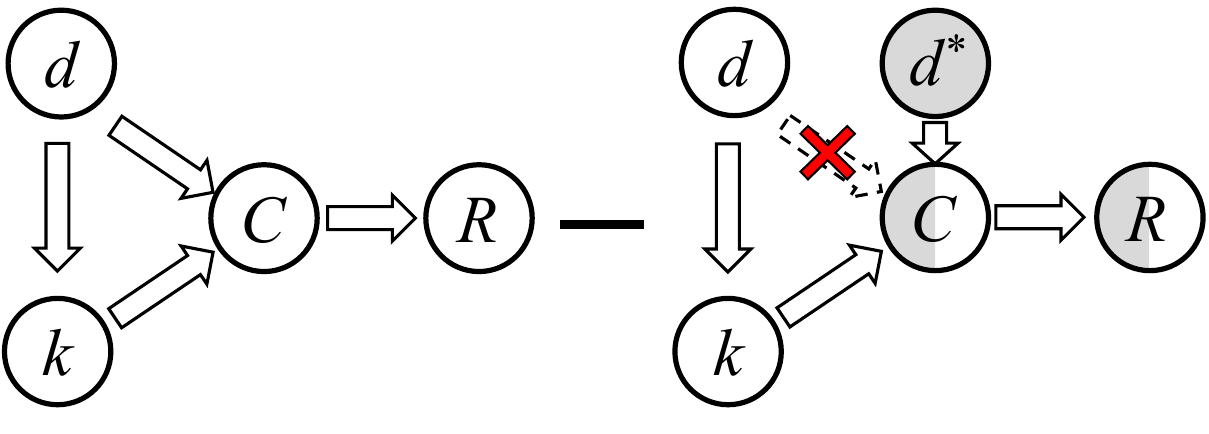}}\\
	\caption{(a) Casual graph for KGD; (b) Comparison between conventional KGD (left) and counterfactual KGD (right). White nodes are at the value $D = d$ and $K = k$ while gray one is $D = d^{*}$. Node $C$, $R$ are counterfactual.}
	\label{fig:causal}
\end{figure}

Except for the pettish demand on stronger knowledge collection techniques, we hope to explore whether there is another direction for alleviating the hallucination of KGD with noise-tolerant methods with the dialogue itself. In this paper, we invoke the insights from \emph{\textbf{counterfactual reasoning}} and \emph{\textbf{causal inference}}~\cite{pearl2000models,pearl2001direct} to investigate this question. After formulating the structural causal model of KGD, we analyze how the generated responses are causally affected by the input elements. As Figure \ref{fig:causal} shows, we introduce two scenarios, conventional KGD and counterfactual KGD, which can be defined as follows:

\noindent \textbf{Conventional KGD}: \emph{What will the response $R$ be, if machine is given the dialogue $D$, and external knowledge $K$?}

\noindent \textbf{Counterfactual KGD}: \emph{What $R$ would the machine say, if it only knows $K$, but is not in a chat as $D$?}

The assumed counterfactual KGD depicts the scenario where the direct effect of dialogue $D$ is reduced. In this case, we can estimate how the external knowledge itself activates the generative model. By subtracting it from the conventional KGD, we naturally utilize the interaction between dialogue and the external knowledge to eliminate the redundant or even mismatched knowledge information from the generation, which enhances the total direct effect of $D$ on responses. This method only modifies the model inference phase, requiring no training stage, thus may be appropriately implemented in different generative models. Based on this analysis, we propose a possible solution with a counterfactual dual-decoding mechanism to alleviate hallucination without training and associated extra data for annotation. 
We conduct an example implementation in Chinese language models to primarily validate our suspicions. For evaluation, we conduct a series of human and automatic evaluation on a public knowledge-grounded dialogue dataset~\cite{zhou2020kdconv} to measure the features of the proposed method in several dialogue metrics. Experimental results prove our idea can indeed mitigate the hallucination without significantly compromising dialogue quality and generation efficiency while keeping able to adapt to different generation models.

The main contribution of this paper is threefold. First, to our best knowledge, our counterfactual reasoning framework is the first to formulate knowledge bias in KGD as causal effects. Second, we provide a causality-based interpretation for recent anti-hallucination KGD works. Third, we propose an alternative direction with counterfactual decoding, which is suitable for different generation models.

\section{Structural Causal Model of KGD}
\label{sec:SCM}

How are the generated responses affected by the input elements? Can we employ the existing information for further exploiting external knowledge? In this section, we formally construct the structural causal model of the Knowledge-grounded Dialogue Generation (KGD) task to answer such questions, which also provides insights into the directions of alleviating the knowledge hallucination from a causal view. 

Structural Causal Model (SCM) is a fundamental tool of causal inference, which formally describes the interactions among the causal features of a certain task~\cite{pearl2000models,pearl2001direct,robinssemantics}. To explore the causal story behind the hallucination phenomenon of this task, we construct its SCM in two stages: 1) formulating the causal graph of KGD task, thereby clarifying the information flow of the general paradigm of it; 2) detecting the potential shortcut of conventional solutions with causal effect analysis. 

\subsection{Task Formulation as Causal Graph}

\emph{Causal Graph}~\cite{pearl2000models} is a directed acyclic graph $\mathcal{G} = \left \{ \mathcal{N}, \mathcal{E} \right \}$, which indicates how a set of variables $\mathcal{N}$ interact with each others through the causal relational links $\mathcal{E}$, \emph{e.g.}, $ X \rightarrow Y$ indicates that variable $Y$ can only obtain value by its prerequisite variable $X$. 

The causal graph of the task of KGD is illustrated in Figure \ref{fig:causal_a}, which is highly general and applicable to a variety of KGD models~\cite{zhou2018commonsense,ghazvininejad2018knowledge,gopalakrishnan2019topical}. Utilizing the language of nodes and links, we study the existing model formulations as:

\textbf{Node $\bm{D}$ (Input Dialogue History).} Dialogue History is a set of conversational utterances between two speakers, formally denoted as $D= \left \{ U_1, S_1, ..., U_{t-1}, S_{t-1}, U_{t} \right \}$, where $U_i$ and $S_i$ are sentences made of words, belonging to the user and the dialogue system respectively. Especially, $U_t$ from the user is also called the \emph{Query}.

\textbf{Node $\bm{K}$ (Retrieved External Knowledge).} External knowledge pool contains multiple pieces of information associated with the dialogue topics, which is denoted as $K=\left \{ k_i \right \}_{i=1}^{|\mathcal{K}|}$, where $k_i$ is a piece of knowledge information. Such knowledge is injected into the dialogue models~\cite{zheng2021exploring} to add extra information. For most of the existing KGD methods, only the most relevant and beneficial knowledge is needed for subsequent processes.

\textbf{Link $\bm{D \rightarrow K}$ (Knowledge Selector).} This stage is commonly defined as \emph{Knowledge Selection}~\cite{zheng2021exploring}, which retrieves the most appropriate external knowledge based on the current dialogue state as:
\begin{equation}
    K := f_{K}(D)
\end{equation}
where node $K$ is causally formed by $D$, and function $f_{K}$ refers to the knowledge selection stage. 

\textbf{Node $\bm{C}$ (Dialogue Context).} This is an intermediate variable of KGD. The dialogue context $C$ is the direct input of the response generation module, which is formed by the combination of dialogue history and selected knowledge via techniques such as encoding integration~\cite{ghazvininejad2018knowledge} or embedding concatenation~\cite{zhou2018commonsense}.

\textbf{Link $\bm{D \rightarrow C}$ and $\bm{K \rightarrow C}$ (Contextual Feature Builder).} The procedure of the joint modeling of the feature of dialogue $D$ and knowledge $K$ to construct dialogue context feature $C$ as:
\begin{equation}
    C := f_{C}(D,K)
\end{equation}
where $C$ contains the direct causal effect from both dialogue history and given knowledge, and function $f_{C}$ refers to the dialogue context modeling stage.

\textbf{Node $\bm{R}$ (Response).} The response $R$ corresponds to the content for answering the current dialogue query $U_t$, which is the target output of the whole KGD task. 

\textbf{Link $\bm{C \rightarrow R}$ (Response Generator).} The procedure of generating the response according to input context, mostly performs as a decoding process of dialogue models~\cite{zhou2018commonsense} or pretrained language models~\cite{bao2021plato},
\begin{equation}
    R := f_{R}(C)
\end{equation}
where function $f_{R}$ refers to the generative process, and response $R$ is caused by the input context $C$.

Therefore, Knowledge-grounded Dialogue (KGD) Generation can be formally defined as: given the dialogue history $D$, the target of this task is to generate a response $R$ for the $t$-th round query $U_t$ with the help of the external knowledge $K$. Once the generated $R$ contains information that conflicts with facts, it is considered to be a \textit{\textbf{hallucinated}} response~\cite{dziri2021neural}.

\begin{figure*}[t]
    \centering
    \includegraphics[width=1.0\linewidth]{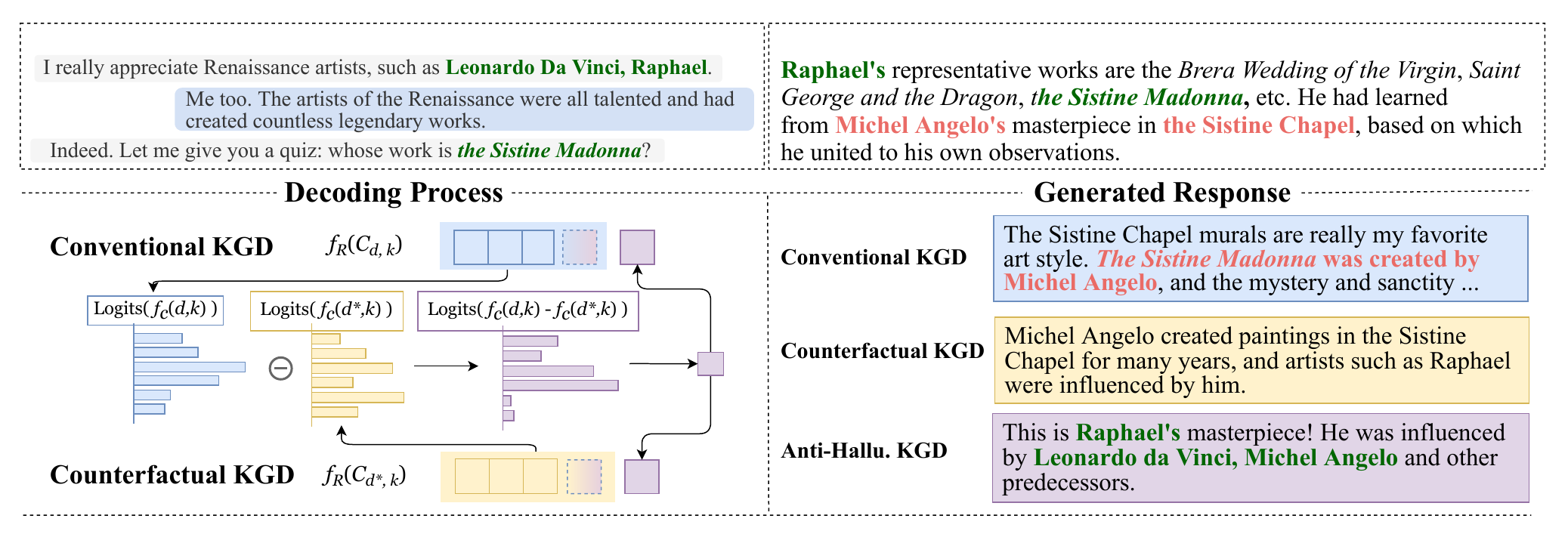}
    \caption{An illustration of counterfactual decoding. Anti-Hallu. is abbreviated from Anti-Hallucination. This mechanism encourages to improve the TDE of the dialogue $D$. The final generated response is more effected by dialogue contexts.}
    \label{fig:frame}
\end{figure*}
\subsection{Causal Effect Analysis}
\label{sec:causal_effect}

The variables of the KGD task naturally follow the above causal dependencies during the whole training and inference processes. However, the conventional KGD models can only observe the output generation results $R$ of the entire graph when given the input $D$ and $K$, lacking the understanding of how these elements affect the final generated response. Fortunately, causal inference~\cite{pearl2018book} provides analytical tools for opening black-box models, with which we can directly manipulate the values of several nodes and conduct effect analysis. Next, we formally introduce these causal tools and then employ them to analyze the shortcut of the current KGD paradigm.

\textbf{Counterfactual} and \textbf{\emph{do}-operation.} \emph{Counterfactual} means ``contrary to the facts''~\cite{roese1997counterfactual}, which is proposed to analyze the role of variables by assigning them hypothetical values. Such thinking is grounded with the approach \textbf{\emph{do}-operation}, denoted as $\bm{do(\cdot)}$. It cuts off the in-coming links of a variable and requires it to take a certain dummy value, \emph{e.g.}, $do(D = d^{*})$ in Figure \ref{fig:causal_b} represents that dialogue is set to a hypothetical value $d^{*}$.
We use the following notations to translate causal graphs into formulas:
\begin{align}
\begin{split}
    C_{d^{*}, K_{d^{*}}} &= f_C\big(do(D = d^{*}), f_K(do(d = d^{*}))\big) \\
    C_{d, k} &= f_C\big(d, f_K(d)\big) \\
    C_{d^{*}, k} &= f_C\big(do(D = d^{*}), f_K(d)\big)
\end{split}
\end{align}

\textbf{Causal Effect.}
This term is broadly defined as a contrast between two \textit{potential outcomes} of the same individual variable under two different \textit{treatments}.
It usually involves the comparison between factual and counterfactual scenarios~\cite{rubin1978bayesian,robins1986new}, \emph{i.e.}, \emph{``What if I do ..., compared with I had done ...''}, and thereby conduct a causal-aware mediation analysis.
Note that causal effect contains a variety of types~\cite{robins1992identifiability,pearl2001direct,vanderweele2013three}, and they are selected appropriately by researchers according to the purpose~\cite{nguyen2021clarifying}.

In the SCM for KGD (as shown in Figure~\ref{fig:causal_a}), dialog history $D$ serves as the treatment, while knowledge $K$ plays the role of the mediator, with the response $R$ as the potential outcome.
The \textit{total effect} (TE) of $D = d$ on the response $R$ compares between two situations with and without the dialog history:
\begin{equation}
    \text{TE} = f_R(C_{d, k}) - f_R(C_{d^{*}, K_{d^{*}}})
\end{equation}
As suggested by \citet{robins1992identifiability}, TE can be further decomposed into \textit{total direct effect} (TDE), which captures the effect of $D$ to the response $R$ flows directly via $D \rightarrow C \rightarrow R$, and \textit{pure indirect effect} (PIE) which depicts the effect indirectly through the mediator $K$ via $D \rightarrow K \rightarrow C \rightarrow R$. Formally:
\small
\begin{equation*}
    \text{TE} = \underbrace{f_{R}\left(C_{d, k}\right) - f_{R}(C_{d^{*}, k})}_{\text{TDE}} + \underbrace{f_{R}\left(C_{d^{*}, k}\right) - f_{R}(C_{d^{*}, K_{d^{*}}})}_{\text{PIE}}
         \end{equation*}
\normalsize

To alleviate the hallucination invoked by mismatched knowledge and knowledge with redundant information, we need to choose a response with an appropriate maximum causal effect considering that hallucination resides in the mediator $K$.
The rationale for choosing a suitable causal effect expresses two principles.
1) To prone possible hallucinated responses that are off the topic of the dialogue history, the causal effect must highlight the influence of the dialog history, avoiding solely over-reliance on the mediate knowledge.
2) To keep a plausible knowledgeable response that contains rich supplemented information, the causal effect must be considered under the presence of the supplemented knowledge and pick up the interaction between $K$ and $D$.

The above two requirements are exactly captured by the \textit{Total Direct Effect} (TDE).
In KGD, it reflects the effect of the dialog history on the response if the knowledge is controlled at the level it would take in the presence of the dialog history.
It is worth noting that, \citet{robins1992identifiability} point out that TE can also be decomposed into \textit{pure direct effect} (PDE) and \textit{total indirect effect} (TIE).
We choose the direct effect rather than the indirect effect because we want to highlight the importance of the dialog history rather than neglecting the dialog history and generate the response with knowledge only.
The reason why we choose TDE rather than PDE is that TDE preserves knowledge in the presence of the dialog history to pick up the interaction between the knowledge and the dialog history to facilitate inter-evaluation.
We show more details on why other causal effects are not suitable for suppressing hallucination in KGD in the Appendix.

Measuring TDE follows $2$ steps:
1) Building the counterfactual scenario where the dialog context $C$ is constructed without dialogue $D$, \emph{i.e.}, $d^{*} = \mathtt{Null}$;
2) Removing the impact of this dialogue-irrelevant knowledge parts based on factual results, denoted as:
$C_{d, k}$ follow the original dialog generation procedure and $C_{d^{*}, k}$ is from the counterfactual, as illustrated in Figure \ref{fig:causal_b}.

In summary, after analyzing the causal effect of the KGD, we invoke the Total Direct Effect:
\begin{equation}    
\text{TDE} = f_{R}\left(C_{d, k}\right) - f_{R}(C_{d^{*}, k})
\end{equation}
to allow the model to use dialog history to verify the correctness and usefulness of retrieved knowledge.
Such analysis encourages us to conduct an appropriate subtraction with a counterfactual scenario ($D = d^{*}$), which provides enlightening insights for the upcoming technical exploration.

\section{Implementation}

Since the optimization direction has been analyzed in a causal view, how to pragmatically implement such TDE in improving the existing KGD paradigm arises as a new question. In this section, we propose a simple but effective approach that performs the required subtraction operation of TDE during the model decoding process. 

Figure \ref{fig:frame} shows a sample of our anti-hallucination decoding mechanism, where the original generation process is synchronized with the counterfactual ($D = d^{*}$), and the subtraction is conducted on the token searching stage. This solution preserves some natural advantages: 1) \emph{Tuning-free}: this method only works on the inference stage, requiring no extra model training; 2) \emph{High-Adaptability}: the parallel processes can be performed with different generative language models.
There are several major technical details during our implementation, including:

\textbf{Knowledge Processing.} Calculating above TDE contains two components. $f_{R}(C_{d, k})$ is from the original generation process while the $f_{R}(C_{d^{*}, k})$ is counterfactual. The dialogue history $D$ is artificially set to almost empty as $d^{*}$. To maintain the comparability of the two processes, we preserve the query of dialogue as the input, which helps the counterfactual process generate an appropriate dialogue-like response.

\textbf{Dual-decoding Strategy.} Let us denote $R = r$ as the final result of the generated response, where $r = \left [ w_{i} \right ]_{1}^{\left | r \right |}$ and $w_{i} \in \mathbb{V}$ is the $i^{th}$ token of it, given the language vocabulary $\mathbb{V}$. For language models~\cite{brown2020language}, when given a probability distribution $p$, the widely-employed approach is to generate text by maximizing the conditional probability. In our decoding process, the $i+1^{th}$ token $w_{i+1}$ of response is searched as follows:
\begin{gather}
     g_{i} =  C_{d,k} \parallel w_{\left [ 1:i \right ]}  \\
     \hat{g}_{i} = C_{d^{*},k} \parallel w_{\left [ 1:i \right ]} \\
     w_{i+1} = \mathrm{argmax} \left ( p\left ( g_{i} \right ) - \lambda (i) \cdot  p\left ( \hat{g}_{i} \right ) \right ) \in \mathbb{V}
\end{gather}
where $\parallel$ is the operation of string concatenation, $w_{\left [ 1:i \right ]}$ is the previously generated response tokens at step $i$. Note that the two processes share the generated tokens. The decay function $\lambda (i)$ is empirically designed to prevent the generation ability from being counteracted, as two processes' inputs are increasingly convergent during decoding.

\section{Experiment}

\begin{table}[t]
\small
\centering
\setlength{\tabcolsep}{4pt}
\begin{tabular}{c|rrr}
\toprule
Annotation                 & Participants  & Label  & Avg-length\\ \midrule
Hallucination        & 100                & 75,000 & 12.8 \\
Human-Chatbot        & 50/50                  & 18,000  & 9.6 \\ \bottomrule
\end{tabular}
\caption{The statistics of the human evaluation protocol, where Avg-length represents the average of utterance length. Hallucination specifically refers to the hallucination annotation on the KdConv generated results.}
\label{tab:human}
\end{table}

\subsection{Experimental Setting}
\label{sec:experimental_setting}

In this section, we conduct a series of evaluations to check whether our analysis can indeed help alleviate hallucination. Specifically, we perform fine-grained hallucination annotation according to~\cite{dziri2021evaluating} and host online tests to analyze the practical performance of our method. Moreover, we also conduct several automatic benchmark evaluations to observe whether our implementation will be reduced significantly in other dialogue metrics.

\textbf{Baseline Language Models.} We utilize several recent typical pretrained models on this benchmark for comparison, including pretrained general language models and dialogue-specific models. The selected baselines include: 
\begin{itemize}
    \item \textbf{CPM-2}~\cite{zhang2021cpm} is the first GPT-architecture Chinese language model, and we employ its $11$B version dense model and adapt it to KGD task.
    \item \textbf{CDial-GPT}~\cite{wang2020large} is a pretrained dialogue model, trained on LCCC conversation corpus with $95.5$M parameters.
    \item \textbf{EVA}~\cite{zhou2021eva} is an excellent $2.8$B parameter Chinese dialogue model based on WDC-dialogue dataset.
    \item \textbf{PLATO-KAG}\footnote{\url{https://github.com/sserdoubleh/plato/tree/develop/projects/PLATO-KAG}. But the public version is only for English datasets WoW~\cite{dinan2018wizard} and Holl-E~\cite{DBLP:conf/emnlp/MogheABK18}. Therefore we translate the KdConv dataset into English and conduct an adaptation on it. The results is for reference only.\label{foot:plato}}. \citeauthor{huang2021plato} propose a joint modeling strategy for the KGD task, upon the PLATO-XL~\cite{bao2021plato}. For online tests, we employ the Xlore2~\cite{jin2019xlore2} as the external knowledge base. All the methods are deployed with XDAI 
toolkit~\cite{yu2022xdai} for adapting to the KGD task.
\end{itemize}

\textbf{Testbed Dataset.}
As our implementation is in Chinese, we employ KdConv~\cite{zhou2020kdconv}, a large Chinese knowledge-grounded dialogue benchmark instead of the widely-used English WoW dataset~\cite{dinan2018wizard}.
KdConv preserves over $4.5$K conversations and $86$K utterances from three topics: Film, Music, and Travel.

\textbf{Model Setup.} Our implementation mainly employs the Chinese-versioned GLM with $10$B parameters~\cite{du-etal-2022-glm} as the backbone dialogue generation model, which is trained on $302$GB raw Chinese data collected from multiple Chinese websites. And the $2.9$B pretrained Transformer-XL (CTXL)~\cite{dai2019transformer} is also utilized for comparison.
The decay function is implemented as $\lambda \left ( i \right ) = \alpha ^{i-1}$, where $\alpha$ is set to $0.3$. The model is deployed with a server of $8$ Nvidia V$100$ GPUs, Intel CPU cores, and $376$GB Memory. 

\textbf{Human Evaluation Protocol.} Previous studies~\cite{liu2016not,chen2017survey} have empirically revealed that human judgments are still the most reliable for evaluating dialogue systems. In this paper, we mainly employ human evaluation for both hallucination annotation and dialogue quality evaluation.
We recruit $100$ people, mostly university students to build our annotation team. For collecting the user online feedback, we deploy the dialogue service as a public demo with the toolkit WeChaty\footnote{\url{https://github.com/wechaty/wechaty}}. Moreover, we develop an online evaluation platform for convenient annotations. Annotators can chat with the online service, label the utterances for different tasks, easily log in and log out, change their labels for already completed problems, or continue evaluation from their current positions freely during the mission period. 

To guarantee the annotation quality, the evaluators are divided into groups of three members, and each group is assigned the same dialogues in certain annotation tasks. Meanwhile, the scores given by any annotators are invisible to others. We also ensure fairness by restricting the topics and initial rounds of conversation uniformly during annotation. Table \ref{tab:human} presents the overall statistics of our annotation.

\subsection{Result Analysis}

Here we analyze a series of experimental results, which indicates the features of this anti-hallucination idea.

\textbf{Hallucination Annotation.} Each method generates five responses on the given dialogue session. For each generated response, three annotators are required to classify it into three categories: \textbf{Factual↑}, \textbf{Spurious↓} and \textbf{Generic}, corresponding to the ratio of the scenario where the generated content is factual, spurious, or not knowledgeable. In general, users hope a KGD system can generate more factual responses. The system may reply a generic perfunctory, but the responses with spurious information are resolutely resisted. Table \ref{tab:hallu} presents the ratios of each model's generated responses in terms of hallucination, where more factual and less spurious labels indicate better performance in alleviating hallucination. The presented values are after the significance test, which provides two main observations: 

(1) Our proposed method can indeed help alleviate the hallucination. After adding the method, GLM's hallucinated responses sharply reduce. One of the most interesting results is that the ``Generic'' ratio is not significantly lifted, which indicates that the rationale behind the improvement is less about erasing all information and more about enabling the model to activate the correct knowledge.

(2) The counterfactual decoding strategy is able to adapt to different generation models. Whether on the main implementation model GLM or the comparison model CTXL, the hallucinations are effectively alleviated. We believe it is promising to further optimize the performance and adapt this thinking to other advanced language models.

\begin{table}[t]
\centering
\setlength{\tabcolsep}{3pt}
\scalebox{0.98}{
\begin{tabular}{l|ccc}
\toprule
\multicolumn{1}{l|}{\multirow{2}{*}{\textbf{Method}}} & \multicolumn{3}{c}{\textbf{Hallucination (\%)}} \\ \cmidrule(l){2-4} 
\multicolumn{1}{c|}{}                        & \textbf{Factual↑}   & \textbf{Generic}   & \textbf{Spurious↓}   \\ \cmidrule(r){1-4}
CPM                                         &    28.57      &      30.71     &    40.71     \\
CDial-GPT                                   &   3.60        &      60.43     & 35.97     \\
EVA                                         &    11.35       &    64.54       &     24.11     \\
PLATO-KAG\textsuperscript{\ref{foot:plato}}     &   10.34      &     8.28      &   81.38       \\ \midrule
CTXL w/o AH                                  &      23.86       &    26.40       &   49.75 \\
CTXL w/ AH                                   &      \textbf{36.04}       &    33.50       &   \textbf{30.45}       \\ \midrule
GLM w/o AH                                  &     46.11       &   20.21        &   38.95 \\
GLM w/ AH                                   &     \textbf{61.65}          &    21.02       &    \textbf{17.33}   \\ \bottomrule
\end{tabular}
}
\caption{The hallucination annotation results of the comparison methods. The values are the ratios of Factual, Generic, and Spurious utterances. ``w/o AH'' and ``w/ AH'' correspondingly represent the settings of without and with our anti-hallucination decoding mechanism.}
\label{tab:hallu}
\end{table}

\begin{figure}[t]
    \centering
    \includegraphics[width=1.01\linewidth]{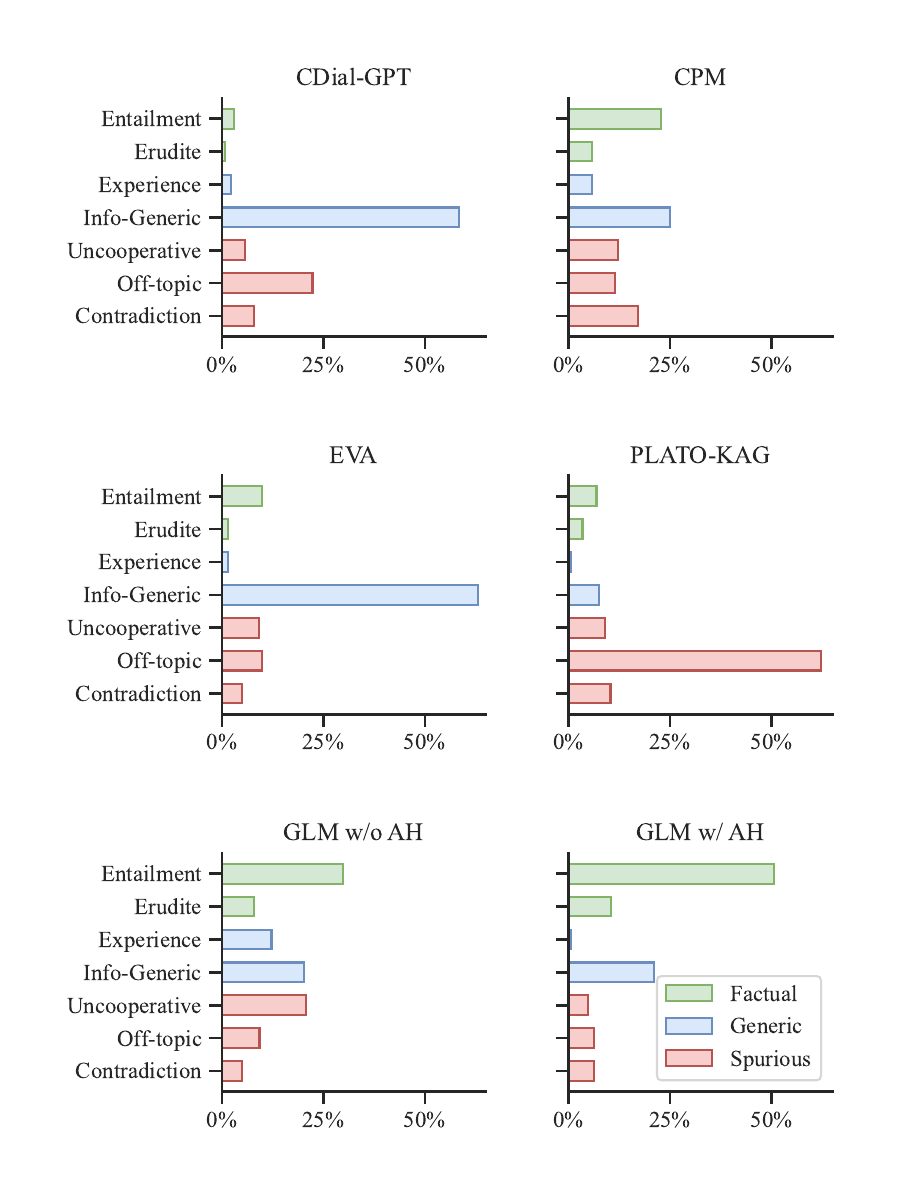}
    \caption{Comparison of the distribution of fine-grained hallucination labels of all comparison methods. The horizontal axis represents the percentage of hallucination labels.}
    \label{fig:fine-grained_hallu}
\end{figure}

\textbf{Fine-grained Annotation.} To further investigate how the hallucinations are alleviated, we follow the insights of~\cite{dziri2021evaluating} to conduct fine-grained hallucination annotation, which requires each utterance to be annotated to a taxonomy\footnote{Detailed introduction and corresponding utterance examples of the fine-grained hallucination labels are in Appendix.}, 
including \textit{Entailment}, \textit{Info-Generic}, \textit{Off-Topic}, and \textit{Contradiction}.
According to the discussed future directions in the paper, we further add \textit{Experience}, \textit{Erudite}, and \textit{Uncooperative} labels.
Note that \textit{Entailment} and \textit{Erudite} are both factual.
The former is explicitly supported by the given knowledge, while the latter is not directly supported but actually correct.
\textit{Uncooperative} and \textit{Experience} are highly related to the dialogue scenario.
We require $100$ annotators to annotate the utterances based on these tags, and each result is also voted by $3$ annotators.

The results shown in Figure \ref{fig:fine-grained_hallu} reveal some details of this anti-hallucination mechanism. (1) For all methods, the factual responses are more \emph{Entailment} than \emph{Erudite}. The former is explicitly supported by the given knowledge, while the latter is not directly supported but correct. This shows that most of the factual results indeed require a high-quality knowledge context, indicating the reason why existing methods~\cite{shuster2021retrieval} can help reduce hallucination. (2) For the comparison within GLM implementation, we find the proportion of responses supported by the given knowledge (\emph{Entailment}) significantly increases, while some of them might be the previous \emph{Off-topic} responses. A plausible explanation is that counterfactual decoding makes the model concentrate on the relevant parts of $K$.

\begin{table}[t]
\centering
\begin{tabular}{l|rr|r}
\toprule
Metric          & w/o AH & w/ AH   & $\Delta$ (\%)   \\ \midrule
Coherence       & 1.803     & 1.795   &   -0.4    \\
Inconsistency↓  & 0.137     & \textbf{0.095} &  \textbf{-30.6}\\
Informativeness & 1.741     & 1.721     &   -1.1   \\
Hallucination↓  & 0.226     & \textbf{0.191} & \textbf{-15.4} \\ \bottomrule
\end{tabular}
\caption{The results of human-chatbot evaluation. The final results presented in the table are the average value of the GLM utterances after filtering out invalid labels.}
\label{tab:human_evaluation}
\end{table}

\begin{table}[t]
\centering
\begin{tabular}{c|rr|rr}
\toprule
\multirow{2}{*}{Setting} & \multicolumn{2}{c|}{Round} & \multicolumn{2}{c}{Duration(s)} \\ \cmidrule(l){2-5} 
                        & Avg         & Max        & Avg         & Max        \\ \midrule
GLM w/o AH                     & 30.5      & 34         & 619.9       & 715.4      \\
GLM w/ AH               & \textbf{32.3}        & \textbf{38}         & \textbf{624.6}       & \textbf{775.2}      \\ \bottomrule
\end{tabular}
\caption{The statistical results of online A/B Test.}
\label{tab:AB_test}
\end{table}

\begin{figure}[t]
    \centering
    \includegraphics[width=1.0\linewidth]{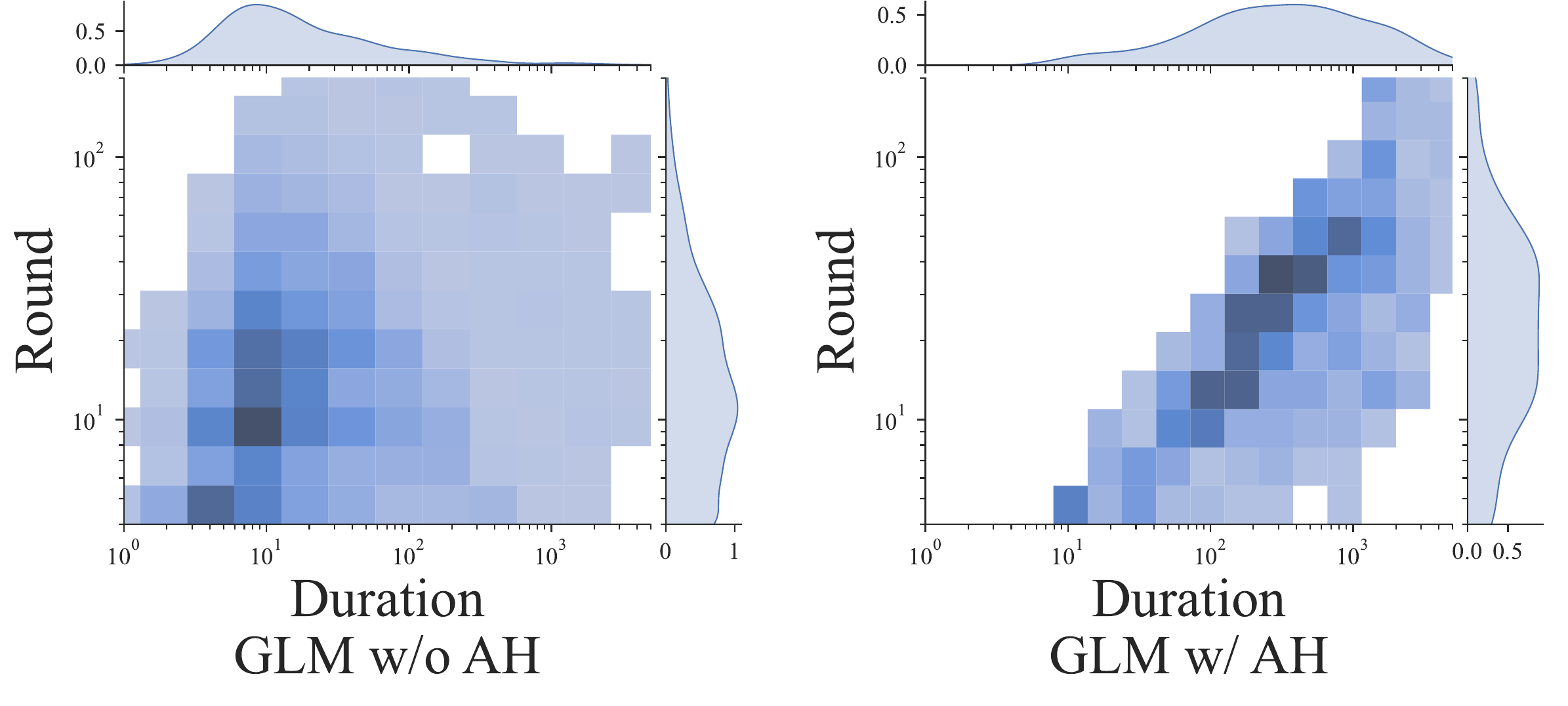}
    \caption{The comparison of the distribution on user involvement distribution of GLM w/ and w/o AH mechanism.
    We show the number of conversation rounds (y-axis) and the total time involved in the conversation (x-axis).
    }
    \label{fig:user}
\end{figure}

\textbf{Human-Chatbot Evaluation.} To investigate the performance of dialogue quality, we divide the annotation team into two $50$-member groups. One group is required to chat with GLM w/ and w/o AH for generating dialogues, while the other group aims to annotate the utterances produced by the prior one.
We follow existing dialogue researches~\cite{bao2021plato,huang2021plato} to employs metrics. \emph{Coherence}: whether the response is consistent with the context, \emph{Informativeness}: whether the response is informative, \emph{Inconsistency}↓: whether the response has conflicts with the dialogue context, \emph{Hallucination}↓: whether the response contains any factual errors (this metric is generally employed and can also doubly check the performance of the previous annotation). Note that the Coherence and Informativeness scale is $[0,1,2]$, whose higher score indicates a better performance while the scale of Inconsistency and Hallucination is $[0,1]$, whose lower score is better. 

According to the evaluation results in Table \ref{tab:human_evaluation}, we find that after counterfactual decoding, the Coherence and Informativeness keep a high performance (with a slight decline of $0.4$\% and $1.1$\%), while Inconsistency↓ drop significantly ($-30.6$\%). As many previous dialogue context contain factual contents, the alleviating of hallucination can also help reducing the Inconsistency. The drop of Hallucination↓ ($-15.4$\%) doubly proves the effectiveness of our strategy on reducing factual errors.
These results can further show that our approach can indeed alleviate the hallucination without significantly destroying the model's other performance.

\textbf{Online A/B Test \& User Involvement Analysis.} We deploy the GLM with and without anti-hallucination decoding in an online environment to collect feedback from users. This evaluation follows the setting of the double-blind A/B test~\cite{man1999double}. Table \ref{tab:AB_test} and Figure \ref{fig:user} shows the collected results of over $300$ participants from May $18$th, 2022 to May $25$th, 2022.

From the statistics, GLM w/ AH outperforms the GLM w/o AH both in terms of dialogue round and duration, especially the maximum of these cases. From the distribution view, the distribution of user activities for the model with the AH strategy is more skewed towards sustaining longer conversations. 
It is worth noting that over $52.1$\% of users chat with GLM w/ AH for more than $10$ rounds. 
These records indicate that: 
(1) Our dual-decoding strategy does not harm the generation quality of the dialogue, such as the conversation engagingness and fluency, which is reflected by the high engagement of real users; (2) GLM w/ AH, that is more ``faithful'' with less hallucinated responses, is more competitive on establishing long-term trust with human beings.

\section{Related Work}

This work is highly relevant with three mainstream topics. They include:

\textbf{Knowledge-grounded Dialogue.} Origin from open-domain dialogue~\cite{ma2020survey}, the task of knowledge-grounded dialogue (KGD) generation aims to generate more informative and faithful responses with the help of various external knowledge~\cite{ghazvininejad2018knowledge}, such as knowledge graphs~\cite{dinan2018wizard}, knowledge documents~\cite{yao2023korc,yu2023kola}, and persona descriptions~\cite{zhang2018personalizing}. With the prosperity of pretrained large language models (LLMs)~\cite{brown2020language,du-etal-2022-glm,bao2021plato}, recent KGD efforts preferably try to employ prompting~\cite{zheng2021exploring,yu2022xdai} or fine-tuning~\cite{xu2021retrieval,zhang2023glm} for exploiting these models in this task. The problem of hallucination in KGD recently attracts the attention of researchers~\cite{dziri2021evaluating}.
However, as the features of LLMs are still being investigated~\cite{liu2021gpt}, a fine-grained KGD result analysis is also essential~\cite{roller2020recipes} to guide the solution of hallucination.

\textbf{Hallucination in NLG.} Factual errors in generative models have been spotted for some time~\cite{holtzman2019curious}, and this topic has become conspicuous along with the rising of LLMs~\cite{petroni2019language}. The problem of \emph{hallucination} (also termed as \emph{unfaithfulness}, \emph{factual consistency}, etc.) widely exists in NLG tasks including dialogue~\cite{santhanam2021rome}, summarization~\cite{pagnoni2021understanding}, translation~\cite{zhou2020detecting} and data2text~\cite{wiseman2017challenges}. Whatever the intrinsic or extrinsic hallucination, their causes are commonly summarized as \emph{heuristic data collection}, \emph{innate divergence} and \emph{knowledge bias}, inspiring the mitigation methods to conduct data filtering and fact-checking for alleviation. Considering the requirement of external sources of these methods, we hope to exploit the dialogue itself in anti-hallucination KGD.

\textbf{Causal Inference.} Counterfactual thinking and causal inference~\cite{pearl2000models,pearl2018book} have inspired several studies in artificial intelligence. Besides its contributions on Computer Vision~\cite{goyal2019counterfactual,niu2021counterfactual}, its application on linguistic tasks~\cite{zmigrod2019counterfactual} is also a vigorous topic, especially data augmentation~\cite{chen2021reinforced} and model explanation~\cite{wu2021polyjuice}. In this paper, we employ this thinking as a theoretical tool for analyzing existing KGD methods and discussing future directions
Since there is no uniform guidance on causal methods in relevant fields, we hope that our efforts can call for more attention to such insights.

\section{Conclusion and Future Work}

In this paper, we study the causal story of alleviating hallucination in knowledge-grounded dialogue generation. After constructing the causal graph of the KGD task and analyzing how the generation results are affected by inputs, we invoke the total direct effect (TDE) to enhance the dialogue quality by utilizing the interaction of dialogue history and external knowledge. Based on this finding, we propose an alternative solution for existing models, which employs a dual-decoding strategy during model inference. We conduct a series of experiments on the example implementation, and the results indicate that this counterfactual decoding strategy can effectively reduce hallucination while maintaining the good performance of other metrics. 

As a primary exploration, this work preserves some limitations that require future investigations. First, it is emergent to apply this idea to more languages and dialogue models, thereby fully validating its features. Second, building a lightweight and practical method for hallucination in dialogue is still promising, especially employing mechanisms such as human-in-the-loop and continuous learning. Third, we sincerely hope our attempts can inspire more efforts in boosting NLG tasks via counterfactual thinking.

\section*{Acknowledgement}

This work is supported by Zhipu AI, a grant from the Institute for Guo Qiang, Tsinghua University (2019GQB0003), and National Natural Science Foundation of China (Grant No.62277034).

\section{Bibliographical References}\label{sec:reference}

\bibliographystyle{lrec-coling2024-natbib}
\bibliography{1-ref}

\end{document}